\documentclass[11pt,a4paper]{article}

\usepackage{algorithm}
\usepackage{algpseudocode}
\usepackage{amsmath}
\usepackage[most]{tcolorbox}
\usepackage[dvipsnames]{xcolor}
\usepackage{hyperref}
\usepackage[all]{hypcap} 
\usepackage{caption}

\usepackage[preprint]{acl}
\usepackage{times}
\usepackage{latexsym}
\usepackage{lineno}
\usepackage[T1]{fontenc}

\usepackage[utf8]{inputenc}

\usepackage{microtype}
\usepackage{booktabs}
\usepackage{multirow}
\usepackage{graphicx}

\usepackage{inconsolata}

\usepackage{graphicx}

\title{LogToPersona: New Benchmark and Efficient Persona Prediction\\ through Semantic Log Filtering}

\author{
\textbf{Ahmed Alajrami\textsuperscript{1}},
 \textbf{Ondrej Bohdal\textsuperscript{1}},
 \textbf{Seokwoo Song\textsuperscript{2}},
 \textbf{Taegeon Um\textsuperscript{2}},\\
 \textbf{Peyman Hosseini\textsuperscript{1}},
 \textbf{Savas Ozkan\textsuperscript{1}},
 \textbf{Mete Ozay\textsuperscript{1}},
 \textbf{Taha Ceritli\textsuperscript{1}}
\\
\\
 \textsuperscript{1}Samsung R\&D Institute UK, United Kingdom,
 \textsuperscript{2}Samsung Research, South Korea
\\
 \small{
   \textbf{Correspondence:} \href{mailto:o.bohdal.1@samsung.com}{o.bohdal.1@samsung.com}
 }
}

\begin{document}
\maketitle
\begin{abstract}
Detailed knowledge of a user's persona is valuable for personalizing generative AI models and agents, yet such details are difficult to obtain and change over time. We address the task of inferring persona attributes from a single day of mobile activity logs. No suitable publicly available benchmark exists for this task, so we introduce a synthetic benchmark of 1{,}600 persona--log pairs. We target on-device inference so that raw logs never leave the phone, which makes memory rather than speed the limiting constraint: a day of activity yields roughly 30K tokens, whose key-value cache alone exceeds the INT4 weights of a 4B-parameter model. We therefore propose a log filtering method that retains only behaviorally distinct entries. Filtering reduces the memory footprint by up to 39\% across nine models from four families and more than halves prefill time, while maintaining or improving prediction quality. For Qwen3-4B it removes 2.5\,GB, bringing the model from well beyond a realistic per-process budget to a manageable one.
Within our benchmark, these results indicate that preserving diverse behavioral events matters more for persona prediction than retaining every semantically similar log entry. Project page: \url{https://aajrami.github.io/LogToPersona}.
\end{abstract}

\section{Introduction}
Knowledge of a user's occupation, routines, preferences, and lifestyle enables models and agents to produce responses that are more relevant and useful~\citep[inter alia]{li2023personalized,salemi2024lamp,tseng-etal-2024-two,Zhang2024PersonalizationOL,Choi2024PICLeED,au2025personalized}. Obtaining such information, however, remains challenging~\citep{Zhang2018PersonalizingDA,Lee2022PERSONACHATGENGP,Chan2024ScalingSD}: users rarely provide comprehensive profiles, and explicitly collected information quickly becomes incomplete or outdated. Behavioral data already available on a user's device offers an alternative source of evidence from which persona attributes may be inferred.

\begin{figure}[!t]
    \centering
    \includegraphics[width=0.99\linewidth]{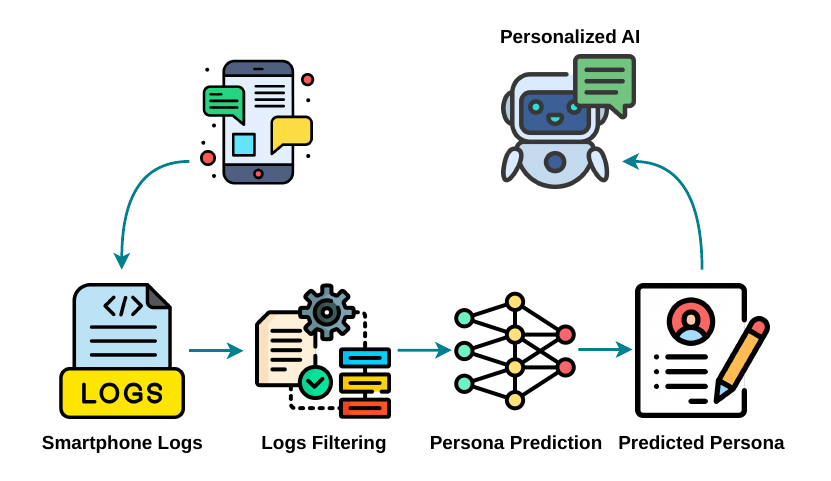}
    \caption{Overview of our persona prediction pipeline with log filtering.}
    \label{fig:overview}
\end{figure}

We study fine-grained persona prediction from one-day mobile activity logs: given a chronological record of a user's actions over 24 hours, the goal is to predict attributes spanning eight categories, from demographics and daily routines to financial behavior and technology habits, yielding a timely persona snapshot without long-term data collection. Figure~\ref{fig:overview} gives an overview of the resulting pipeline.

Progress is limited by the absence of suitable evaluation resources. Public mobile-sensing datasets lack fine-grained annotations across these dimensions, and collecting them at scale would be costly and raise significant privacy concerns. We therefore construct a synthetic benchmark of 1{,}600 detailed personas paired with corresponding 24-hour mobile activity logs.

Privacy is also a central constraint: mobile activity logs may reveal highly sensitive information, and transmitting them to remote servers introduces substantial risk~\citep{huang-etal-2025-logrules,siyan-etal-2025-papillon}. We therefore target on-device inference, in which raw activity data remains local.

On-device deployment is limited by memory before it is limited by speed~\citep{dhar2021survey,Liu2024MobileLLMOS,alizadeh-etal-2024-llm,siddiqui-etal-2025-llms}. Persona prediction is naturally a background task, run when the device is idle, so a prefill cost of seconds is acceptable, but memory has no such flexibility. A model that does not fit is not slow, it is terminated by the operating system, and the feature cannot ship. The binding quantity is the peak resident footprint: model weights plus the key--value (KV) cache, which grows linearly with input length. For a full day of activity, 29.7K tokens on average in our benchmark, the KV cache of Qwen3-4B occupies 4.1\,GB at FP16, 1.8 times the size of the same model's INT4-quantized weights. Weight quantization, the standard on-device remedy, therefore addresses only the smaller of the two terms, and reducing the input is necessary.

A full day's log contains substantial repetition--repeated application launches, similar queries, recurring system events--that lengthens the input but adds little evidence once the underlying behavior has been observed. We therefore propose a semantic similarity-based filtering method that retains behaviorally distinct entries while removing near-duplicates, compressing the average input by 61\%. That prediction quality is preserved under this compression suggests that the diversity of behavioral evidence matters more than the completeness of the record.

Our contributions are as follows:
\begin{itemize}
    \item We formalize fine-grained persona prediction from one-day mobile activity logs under on-device memory and privacy constraints.
    \item We introduce a synthetic benchmark of 1{,}600 persona--log pairs spanning demographic, behavioral, and lifestyle attributes.
    \item We propose a semantic similarity-based log filtering method that reduces the memory footprint by up to 39\% and prefill time by more than 50\% across nine models from four families, while maintaining or improving prediction quality.
\end{itemize}

\section{Related Work}
\paragraph{Persona-Based Personalization.}
Prior work has developed persona representations primarily to support personalized dialogue, role-playing, and synthetic data generation. \citet{Zhang2018PersonalizingDA} crowdsourced 1{,}155 personas, each represented by at least five profile sentences, and used them to construct a personalized dialogue dataset. More recently, \citet{Chan2024ScalingSD} introduced Persona Hub, a collection of one billion personas automatically curated from web data to support large-scale synthetic data generation. \citet{castricato-etal-2025-persona} generated 1{,}586 synthetic personas from U.S. Census data and used them to evaluate language models' ability to role-play diverse users. Similarly, \citet{jandaghi-etal-2024-faithful} proposed Synthetic-Persona-Chat, which expands seed personas into 5{,}000 user profiles and generates more than 20{,}000 persona-conditioned conversations.

Other work has collected persona information directly from participants. \citet{yamashita-etal-2023-realpersonachat}, for example, gathered persona descriptions and personality traits from 233 participants and used them to construct a Japanese conversational corpus. Whereas these studies use predefined personas to condition or evaluate language models, we study the inverse problem: inferring a structured persona from observed user behavior.

\paragraph{Behavioral Inference from Mobile Data.}
Mobile and wearable sensing studies have shown that device activity can provide evidence about user behavior and well-being. In StudentLife, \citet{wang2014studentlife} collected smartphone sensing data from 48 college students over an academic term and examined relationships between behavioral signals and outcomes such as mental health and academic performance. This line of work typically predicts specific outcomes from sensor measurements collected over extended periods. In contrast, we infer a broader set of demographic, lifestyle, routine, and technology-related attributes from a single day of application- and action-level logs. We additionally consider the computational constraints of performing language-model-based inference locally on the device.

\begin{table*}[!t]
\centering
\small
\begin{tabular}{@{}p{0.27\linewidth}p{0.66\linewidth}@{}}
\toprule
Category & Attributes \\
\midrule
Demographics & Age, gender, place of residence, marital or family status \\
Job and Lifestyle & Occupation, work schedule, workplace location, commuting behavior \\
Daily Patterns & Sleep schedule, weekday and weekend routines, hobbies \\
Financial Behavior & Spending habits, investment preferences, credit management \\
Health and Fitness & Exercise routines, general health practices \\
Social Activities & Primary relationships, communication patterns, community engagement \\
Technology Habits & Frequently used applications, device-usage patterns \\
Motivations and Pain Points & Personal goals, recurring concerns, socioeconomic stressors \\
\bottomrule
\end{tabular}
\caption{The eight persona categories and the attributes recorded in each.}
\label{tab:persona-structure}
\end{table*}

\paragraph{Context and Prompt Compression.}
Reducing long language-model inputs has been studied through prompt compression and token-selection methods. LLMLingua-2~\citep{pan-etal-2024-llmlingua}, for example, uses a token-classification model to identify and remove low-information tokens from an input prompt. Such methods are generally task-agnostic and operate at the token level, which may disrupt the structure of individual mobile-log events. Our method instead exploits the repeated event structure of behavioral logs: it compares complete entries in an embedding space and removes semantically redundant events while preserving representative activities. The objective is therefore not only to shorten the input, but also to retain a diverse set of behavioral evidence for persona inference.

Reducing the resident footprint rather than the input is complementary to our work: sub-billion-parameter architectures~\citep{Liu2024MobileLLMOS}, grouped-query attention~\citep{ainslie-etal-2023-gqa}, and flash-based parameter streaming~\citep{alizadeh-etal-2024-llm} all shrink what needs to be held in memory for a fixed input. These techniques act on the model, whereas our method acts on the input, and the two approaches can be combined.

Overall, existing work has investigated persona construction, behavioral prediction, and prompt compression largely as separate problems. Our work connects these directions by introducing fine-grained persona inference from mobile activity logs and studying how structure-aware log filtering can make this task more efficient for on-device language models.

\section{Benchmark}

The sensitive nature of mobile activity logs makes it difficult to construct and release real-world datasets with detailed persona annotations. Existing mobile-sensing datasets typically contain either coarse behavioral outcomes or a limited set of self-reported attributes, rather than the broad persona representation considered in this work. To support controlled development and evaluation, we introduce a synthetic benchmark containing 1{,}600 persona--log pairs. Each pair includes a structured persona and a corresponding 24-hour mobile activity timeline.

\subsection{Dataset Generation}
We generate the benchmark using \textit{Gemma-4 31B}~\citep{gemmateam2026gemma4technicalreport} with structured prompts designed to produce diverse and internally consistent persona--log pairs. The generation process consists of three stages. First, the model generates a detailed persona spanning the eight categories described in Section~\ref{sub_sec:persona_structure}. Second, conditioned on the complete persona, it generates an activity timeline. Finally, it uses the detailed persona and activity timeline to generate a chronological mobile activity log covering a full 24-hour period.

Generated data are checked for structural completeness, temporal consistency, and agreement between the persona and the corresponding activity log. Samples that contain malformed timelines, contradictory attributes, or implausible activity sequences are regenerated. Appendix~\ref{sec:prompts} provides the complete generation prompts.

\subsection{Persona Structure}
\label{sub_sec:persona_structure}
Each persona is organized into the eight categories listed in Table~\ref{tab:persona-structure}. This representation extends beyond demographic attributes to include behavioral and lifestyle dimensions relevant to personalization. Not every attribute is equally observable from a single day of activity: some are supported directly by logged actions, others only indirectly through behavioral patterns. We therefore treat the benchmark as an evaluation of evidence-based persona estimation rather than assuming that every ground-truth detail is fully recoverable from the logs.

\subsection{Log Structure}
Each sample contains a smartphone activity timeline spanning 24 hours and divided into hourly blocks. Individual entries record the application or service involved, the action performed, a timestamp or time interval, and relevant contextual information. Depending on the event type, the context may include activities such as opening an application, issuing a query, viewing content, communicating with another person, completing a transaction, or triggering a system-level event.

The generated logs are conditioned on the corresponding personas so that their activities plausibly reflect the user's occupation, routines, preferences, and lifestyle. 
The logs are not intended to state every persona attribute explicitly, but instead they provide behavioral evidence from which the attributes may be inferred.

\algrenewcommand\algorithmicrequire{\textbf{Input}}
\algrenewcommand\algorithmicensure{\textbf{Output}}

\begin{algorithm*}[t]
\footnotesize
\centering
\begin{minipage}{0.9\textwidth}
\caption{Semantic Similarity Log Filtering} \label{algo:filtering_pseudocode}
\begin{algorithmic}[1]
\Require Log sequence $L = [l_1, l_2, \ldots, l_n]$, distance threshold $\tau$
\Ensure Filtered log sequence $L'$
\State $E \gets [\text{Embed}(l_1),\; \text{Embed}(l_2),\; \ldots,\; \text{Embed}(l_n)]$ \Comment{Embed all log entries}
\State $\textit{index} \gets \text{FAISS\_Build}(E)$ \Comment{Build FAISS index}
\State $\textit{skip} \gets \emptyset$ \Comment{Set of indices to filter out}
\For{$i = 1$ \textbf{to} $n$}
    \If{$i \in \textit{skip}$}
        \State \textbf{continue}
    \EndIf
    \State $\mathcal{N} \gets \text{FAISS\_Search}(\textit{index},\; E[i])$ \Comment{Retrieve nearest neighbors}
    \For{$(j,\; d) \in \mathcal{N}$}
        \If{$j \neq i$ \textbf{and} $d < \tau$ \textbf{and} $j \notin \textit{skip}$}
            \State $\textit{skip} \gets \textit{skip} \cup \{j\}$ \Comment{Mark neighbor for removal}
        \EndIf
    \EndFor
\EndFor
\State $L' \gets [l_i \;\textbf{for}\; i \in \{1, \ldots, n\} \setminus \textit{skip}]$ \Comment{Retain unskipped entries in order}
\State \Return $L'$
\end{algorithmic}
\end{minipage}
\end{algorithm*}

\subsection{Dataset Splits}
The benchmark contains 1{,}600 persona--log pairs, randomly partitioned into 1{,}000 training samples for fine-tuning, 100 validation samples for hyperparameter selection, including the semantic filtering threshold, and 500 test samples for final evaluation. The splits are constructed to maintain diversity across persona categories and reduce systematic bias in the evaluation.

\section{Log Filtering via Similarity-Based Deduplication}
A full day of mobile activity often contains repeated application launches, similar queries, and recurring system events. Processing all entries increases the language model's input length even when many events provide nearly identical behavioral evidence. We therefore propose a similarity-based deduplication method that retains representative activities while removing semantically redundant entries.

Given a chronological log
$\mathcal{L} = \{l_1, l_2, \ldots, l_n\},$
our method consists of three stages: embedding each log entry, retrieving semantically similar entries, and selecting a temporally ordered subset of representative events.

\subsection{Log Embedding}

Each log entry $l_i$ is independently encoded as a dense vector
$\mathbf{e}_i = f_{\mathrm{emb}}(l_i)$,
where $f_{\mathrm{emb}}$ is the Qwen3-Embedding-0.6B model~\citep{qwen3embedding}. We select this model because its relatively small parameter count is compatible with resource-constrained deployment while still providing semantic representations that distinguish different user activities.

\subsection{Similarity Search}
We construct a FAISS index~\citep{douze2024faiss} over the embeddings $\mathcal{E} = \{\mathbf{e}_1, \mathbf{e}_2, \ldots, \mathbf{e}_n\}.$
We use a \textit{Flat} index with \textit{exact} nearest-neighbor search and $L_2$ distance as the similarity measure. For each entry, the index returns its $k=100$ nearest neighbors, excluding the query entry itself.\footnote{At the log lengths evaluated here, the same result could be obtained by direct pairwise comparison. We use FAISS so that the procedure scales to longer observation windows.}

\subsection{Iterative Deduplication}
The entries are processed sequentially in their original temporal order. We maintain a skip set $\mathcal{S}$, initially empty, containing entries already identified as redundant. Each entry not in $\mathcal{S}$ is retained as a representative event, and every neighbour $l_j$ with $\lVert \mathbf{e}_i - \mathbf{e}_j \rVert_2 < \tau$ is added to $\mathcal{S}$. The retained entries form the filtered log $\widetilde{\mathcal{L}} \subseteq \mathcal{L}$.

This procedure is intentionally conservative at the event level: complete log entries are either retained or removed. Unlike token-level compression, it does not modify the internal text of an event, reducing the risk of separating an action from its application, timestamp, or contextual information.

\subsection{Distance Threshold Selection}
The threshold $\tau$ controls the trade-off between compression and information preservation. A smaller value removes only highly similar entries, whereas a larger value produces greater compression but may merge activities that contain distinct persona evidence.

We select $\tau$ using the validation set. Specifically, we evaluate candidate values from \textit{[0.1, 0.15, 0.2, 0.25, 0.3, 0.5]}, apply Semantic Similarity filtering, and evaluate the resulting logs using both the average persona prediction score and the average language-model prefill time. We then select the threshold that provides the best overall accuracy--efficiency trade-off, favoring configurations that reduce prefill time and memory footprint without degrading persona prediction quality. This procedure yields $\tau = 0.2$.

The selected threshold is fixed for all test-set experiments and all evaluated language models.

\subsection{Compressed Log Representation}
The final output is a chronologically ordered subset of distinct log entries that is provided directly to the persona prediction model. On average, semantic filtering reduces the average input length from \textit{29.7K} to \textit{11.6K} tokens.

The filtering stage requires an average of 1.4 seconds per sample in our experimental setup. In Section~\ref{sec:results}, we evaluate whether this preprocessing cost is offset by lower language-model prefill time and whether the compressed representation preserves persona prediction quality.

Algorithm~\ref{algo:filtering_pseudocode} presents the pseudocode of the semantic similarity log filtering method.

\section{Experimental Setup}
We evaluate persona prediction on the 500 persona--log pairs of the benchmark's test split, comparing filtering methods on both prediction quality and computational efficiency.

\subsection{Models}
We evaluate nine instruction-tuned open-weight models spanning four families and a range of parameter scales: Qwen3~\citep{qwen3technicalreport} (4B, 8B, 14B), Gemma-3~\citep{gemmateam2025gemma3technicalreport} (4B, 12B), Gemma-4~\citep{gemmateam2026gemma4technicalreport} (E2B, E4B), and Ministral-3~\citep{Liu2026Ministral3} (8B, 14B). This lets us assess whether the effects of log filtering generalize across architectures and capacities. All models receive the same persona-prediction instruction, given in Appendix~\ref{sec:persona_prediction_prompts}, only the input log representation varies across conditions.

\renewcommand*{\arraystretch}{0.98}
\begin{table*}[!t]
\begin{center}
\small
\resizebox{0.9\textwidth}{!}{%
\begin{tabular}{lcccc|cc|cc|cc}
\toprule

& & \multicolumn{9}{c}{\textbf{Avg PPS} ($\uparrow$)} \\
\textbf{Method} & \textbf{Avg Log Length} & \multicolumn{3}{c}{\textbf{Qwen3}} & \multicolumn{2}{c}{\textbf{Gemma-3}} & \multicolumn{2}{c}{\textbf{Gemma-4}} & \multicolumn{2}{c}{\textbf{Ministral-3}} \\
& \textbf{(tokens)} & \textbf{4B} & \textbf{8B} & \textbf{14B} & \textbf{4B} & \textbf{12B} & \textbf{E2B} & \textbf{E4B} & \textbf{8B} & \textbf{14B} \\

\midrule

No Filtering & 29.7K & 69.8 & 76.7 & 73.8 & 69.3 & 74.4 & 75.1 & 75.3 & \textbf{76.8} & 77.3 \\ 
LLMLingua-2 & 15.4K & 58.3 & 75.0 & 73.8 & 68.9 & 71.5 & 73.3 & 72.9 & 75.3 & 76.2  \\
50\% RS & 14.8K & 62.1 & 73.8 & 73.1 & 67.6 & 72.2 & 73.4 & 73.2 & 74.1 & 74.0 \\
Exact Match & 13.4K & \textbf{70.0} & 75.3 & \textbf{78.6} & \textbf{69.5} & 76.0 & \textbf{75.4} & \textbf{76.2} & 76.7 & \textbf{78.0} \\
Semantic Similarity & 11.6K & 69.7 & \textbf{76.8} & 76.6 & 69.3 & \textbf{76.1} & 75.0 & 76.0 & 76.4 & 77.5 \\

\bottomrule
\end{tabular}%
}
\caption{Average persona prediction score (PPS) results in zero-shot setting. \textbf{Bold} values denote the best performance across all filtering methods.} 
\label{table:ss_zero_shot_results}
\end{center}
\end{table*}

\subsection{Baselines}

We compare our proposed similarity-based filtering method against the following three baselines:
\begin{itemize}
    \item{\textbf{Full log (no filtering)}}, where the entire 24-hour log is provided as input to the model without any filtering. This represents an upper bound on the information available to the model and serves as the primary baseline, though it incurs the highest computational cost.
    \item{\textbf{50\% Random sampling (RS)}}, where we randomly sample 50\% of the log entries, discarding the remainder. This baseline controls for the effect of simply reducing the input length, without any consideration of the content or diversity of the retained entries.
    \item{\textbf{LLMLingua-2}}~\citep{pan-etal-2024-llmlingua}, a prompt compression method that uses a small language model\footnote{\url{https://huggingface.co/microsoft/llmlingua-2-xlm-roberta-large-meetingbank}}, fine-tuned for token classification to perform task-agnostic prompt compression, to predict the importance of each token and remove low-information tokens. Unlike our approach, which filters at the log-entry level based on semantic similarity, LLMLingua-2 operates at the token level and does not leverage the structure of the log data.

\end{itemize}
In addition, we propose another filtering method using \textbf{exact match}, where we remove log entries that are exact duplicates of previously seen entries, retaining only the first occurrence. This method tests whether string-level deduplication alone is sufficient, as opposed to the semantic similarity criterion employed by our method.

\subsection{Evaluation Metrics}
We evaluate each approach along three dimensions: the quality of the predicted persona, the latency of inference, and its memory footprint.
\begin{itemize}
    \item{\textbf{Persona Prediction Score (PPS):}}
    We score predictions across the eight persona categories in Section~\ref{sub_sec:persona_structure}, weighting categories and attributes within each category equally. Numeric and unambiguous categorical attributes, such as age, use exact match, while free-form attributes, such as occupation and location, use semantic similarity with \textit{intfloat/e5-large-v2}~\citep{wang2022text}. Scores range from 0 to 1 and are multiplied by 100 for reporting on a 0--100 scale. Invalid or unparsable persona outputs receive a score of zero.

    \item{\textbf{Prefill Time:}} The average time, in seconds, spent processing a request's input tokens before autoregressive generation, measured over the test set with the vLLM \citep{kwon2023efficient} metric \texttt{request\_prefill\_time\_seconds} on NVIDIA A40 GPUs, under a configuration held identical across filtering methods (Appendix~\ref{app:implementation}).

    \item{\textbf{Memory Footprint:}}
    We estimate memory as quantized model weights plus the KV cache. For an input of \(n\) tokens, the KV-cache requirement is
\begin{equation*}
K(n)=2b\left(L_g h_g d_g\,n + L_l h_l d_l \min(n,w)\right),
\end{equation*}
where \(b\) is bytes per cached element, \(L_g\) and \(L_l\) are the numbers of global and local attention layers, \(h\) and \(d\) denote KV heads and head dimension, and \(w\) is the local-attention window. For dense attention, this reduces to \(2Lhd\,nb\). The estimate depends only on model architecture and input length, not on the accelerator. It excludes activations, runtime overhead, and decoding-time cache growth, so reported values are lower bounds on peak memory.

\end{itemize}

\section{Results}
\label{sec:results}
We evaluate the proposed log-filtering methods across four model families, Qwen3, Gemma-3, Gemma-4, and Ministral-3, with multiple parameter scales from each family. For every model, we compare five input configurations: (1) the full 24-hour log, (2) LLMLingua-2 compression, (3) 50\% random sampling (RS), (4) Exact Match deduplication, and (5) Semantic Similarity filtering. We evaluate persona prediction quality using the average Persona Prediction Score (PPS), reported in Table~\ref{table:ss_zero_shot_results}, and inference efficiency using average prefill time (Table~\ref{table:prefill_zero_shot_results}) and memory footprint (Table~\ref{tab:memory}).

\renewcommand*{\arraystretch}{0.98}
\begin{table*}[!t]
\begin{center}
\small
\resizebox{0.9\textwidth}{!}{%
\begin{tabular}{lcccc|cc|cc|cc}
\toprule

& & \multicolumn{9}{c}{\textbf{Avg Prefill Time (sec)} ($\downarrow$)} \\
\textbf{Method} & \textbf{Avg Log Length} & \multicolumn{3}{c}{\textbf{Qwen3}} & \multicolumn{2}{c}{\textbf{Gemma-3}} & \multicolumn{2}{c}{\textbf{Gemma-4}} & \multicolumn{2}{c}{\textbf{Ministral-3}} \\
& \textbf{(tokens)} & \textbf{4B} & \textbf{8B} & \textbf{14B} & \textbf{4B} & \textbf{12B} & \textbf{E2B} & \textbf{E4B} & \textbf{8B} & \textbf{14B} \\

\midrule

No Filtering & 29.7K & 12.0 & 17.4 & 17.6 & 3.8 & 11.1 & 5.7 & 11.1 & 7.5 & 8.4 \\
LLMLingua-2 & 15.4K & 10.2 & 9.7 & 10.2 & 2.3 & 6.2 & \textbf{2.6} & 4.8 & \textbf{3.0} & 4.5 \\
50\% RS & 14.8K & 9.3 & 9.4 & 10.4 & 2.2 & 6.3 & 3.3 & 5.7 & 3.6 & 4.0 \\
Exact Match & 13.4K & 6.0 & 7.2 & 8.3 & 2.1 & 5.1 & 2.8 & 5.5 & 3.3 & 4.0 \\
Semantic Similarity & 11.6K & \textbf{4.7} & \textbf{5.7} & \textbf{7.1} & \textbf{1.7} & \textbf{4.5} & 2.8 & \textbf{4.5} & \textbf{3.0} & \textbf{3.5} \\

\bottomrule
\end{tabular}%
}
\caption{Average Prefill Time results in zero-shot setting. \textbf{Bold} values denote the best performance.} 
\label{table:prefill_zero_shot_results}
\end{center}
\end{table*}

\subsection{Persona Prediction Quality}
Table~\ref{table:ss_zero_shot_results} shows that Exact Match and Semantic Similarity filtering generally preserve or improve persona prediction quality across all four model families and every parameter scale, despite discarding 55\% and 61\% of the input respectively. The largest gains appear at the larger scales, where Semantic Similarity improves Qwen3-14B by 2.8 points and Gemma-3-12B by 1.7. Exact Match attains the best score of any configuration for six of the nine models. Where filtering does not help it does not appreciably hurt: the largest degradation under Semantic Similarity is 0.4 points, for Ministral-3-8B.

LLMLingua-2 and 50\% Random Sampling produce the lowest scores throughout, with LLMLingua-2 costing Qwen3-4B 11.5 points relative to the full log. Reducing input length alone is therefore insufficient: effective compression must preserve complete and behaviorally distinct log entries. These results support our hypothesis that repeated entries contribute redundant or distracting information rather than additional signal for persona inference.

\subsection{Inference Efficiency}
Table~\ref{table:prefill_zero_shot_results} reports the average prefill time for each model and filtering configuration. Semantic Similarity filtering reduces prefill time by at least 50\% for every evaluated model, ranging from 51\% for Gemma-4-E2B to 67\% for Qwen3-8B.

Both Semantic Similarity filtering and LLMLingua-2 introduce preprocessing overhead before language-model inference, but Semantic Similarity requires only 1.4 seconds per sample on average, against 1.8 seconds for LLMLingua-2, and this cost is more than offset by the prefill reduction it enables. 

\subsection{Memory Feasibility}
Table~\ref{tab:memory} reports model-weight and KV-cache memory at the full and filtered input lengths, assuming INT4 weight-only quantization and an FP16 KV cache~\citep{lin2023awq,Gunter2024AppleIF}. Accuracy and prefill results are measured at native precision, so the memory analysis assumes that INT4 quantization preserves prediction quality. Filtering affects only the cache term. For example for Qwen3-4B it removes 2.5\,GB of KV memory, reducing the total from 6.3\,GB to 3.8\,GB under INT4 weights.

At the full 29.7K-token input length, KV memory is comparable to or larger than model-weight memory: Qwen3-4B requires 4.1\,GB of KV cache versus 2.2\,GB of quantized weights, while Ministral-3-8B requires 3.9\,GB versus 4.7\,GB. Because weight quantization does not reduce KV memory, long contexts can remain prohibitive even after aggressive weight compression. Under a permissive 4\,GB budget, the full Qwen3-4B configuration exceeds the budget before runtime overhead, whereas filtering brings its weight-plus-KV estimate close to that boundary.

The gains are largest for models with dense attention. Filtering reduces total memory by 22--39\% for Qwen3 and Ministral-3, but only 11--13\% for Gemma-3 and Gemma-4, whose sliding-window and cache-sharing mechanisms already constrain KV growth. Thus, filtering is most complementary to weight-side techniques such as quantization, which leave the KV cache unchanged.

\begin{figure}[!t]
    \centering
    \includegraphics[width=0.98\linewidth]{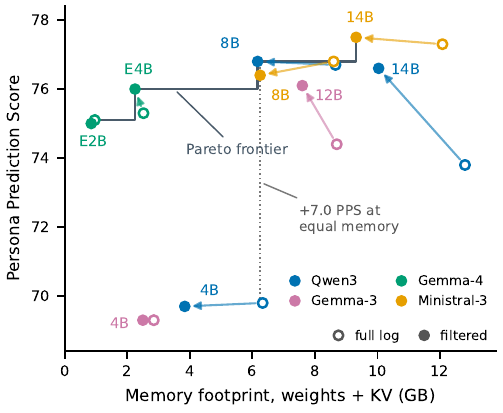}
    \caption{Projected memory–quality operating points under INT4 weight-only quantization, assuming quantization preserves the native-precision PPS reported in Table~\ref{table:ss_zero_shot_results}.
    }
    \label{fig:tradeoff}
\end{figure}

\textbf{Trading context for capacity.}
At nearly the same memory budget, Qwen3-4B with the full log requires 6.3\,GB and reaches 69.8 PPS, whereas Qwen3-8B with filtered logs requires 6.2\,GB and reaches 76.8 PPS. Ministral-3-8B similarly reaches 76.4 PPS at 6.2\,GB (Figure~\ref{fig:tradeoff}). Four of the five points on the memory--quality Pareto frontier use filtered logs, suggesting that under a fixed memory budget, a larger model with compressed context can outperform a smaller model with the complete log.

\renewcommand*{\arraystretch}{0.95}
\begin{table*}[!t]
\centering
\small
\resizebox{0.9\textwidth}{!}{%
\begin{tabular}{l r r r r r r r}
\toprule
& & \multicolumn{2}{c}{No Filtering (29.7K tok)} & \multicolumn{2}{c}{Semantic Sim. (11.6K tok)} & \multicolumn{2}{c}{Reduction} \\
\cmidrule(lr){3-4} \cmidrule(lr){5-6} \cmidrule(lr){7-8}
Model & Weights & KV & Total & KV & Total & KV & Total \\
& (GB) & (GB) & (GB) & (GB) & (GB) & & \\
\midrule
Qwen3-4B & 2.2 & 4.1 & 6.3 & 1.6 & \textbf{3.8} & 61\% & 39\% \\
Qwen3-8B & 4.6 & 4.1 & 8.7 & 1.6 & 6.2 & 61\% & 29\% \\
Qwen3-14B & 8.3 & 4.5 & 12.8 & 1.8 & 10.0 & 61\% & 22\% \\
Gemma-3-4B & 2.2 & 0.7 & \textbf{2.8} & 0.3 & \textbf{2.5} & 51\% & 12\% \\
Gemma-3-12B & 6.6 & 2.1 & 8.7 & 1.0 & 7.6 & 52\% & 13\% \\
Gemma-4-E2B & 0.8 & 0.2 & \textbf{1.0} & 0.1 & \textbf{0.9} & 59\% & 11\% \\
Gemma-4-E4B & 2.0 & 0.5 & \textbf{2.5} & 0.2 & \textbf{2.2} & 58\% & 11\% \\
Ministral-3-8B & 4.7 & 3.9 & 8.6 & 1.5 & 6.2 & 61\% & 27\% \\
Ministral-3-14B & 7.5 & 4.5 & 12.1 & 1.8 & 9.3 & 61\% & 23\% \\
\bottomrule
\end{tabular}
}
\caption{On-device memory footprint (model weights $+$ KV cache) with INT4 weight-only quantization and an FP16 KV cache.
Weights are text-only. Totals are computed before rounding and may differ from the sum of the rounded columns. 
\textbf{Bold} totals fit within a 4\,GB per-process budget.
}
\label{tab:memory}
\end{table*}

\subsection{Fine-tuning Experiments}
To determine whether supervised adaptation provides additional gains, we fine-tune Qwen3-4B using LoRA~\citep{hu2022lora}, taking it as representative of general-purpose dense-attention architectures at a scale plausible for on-device deployment. A separate model is trained for each input configuration, with the same filtering method applied to the training and test sets and hyperparameters held fixed throughout (Appendix~\ref{app:implementation}).

\begin{figure}[!t]
    \centering
    \includegraphics[width=0.98\linewidth]{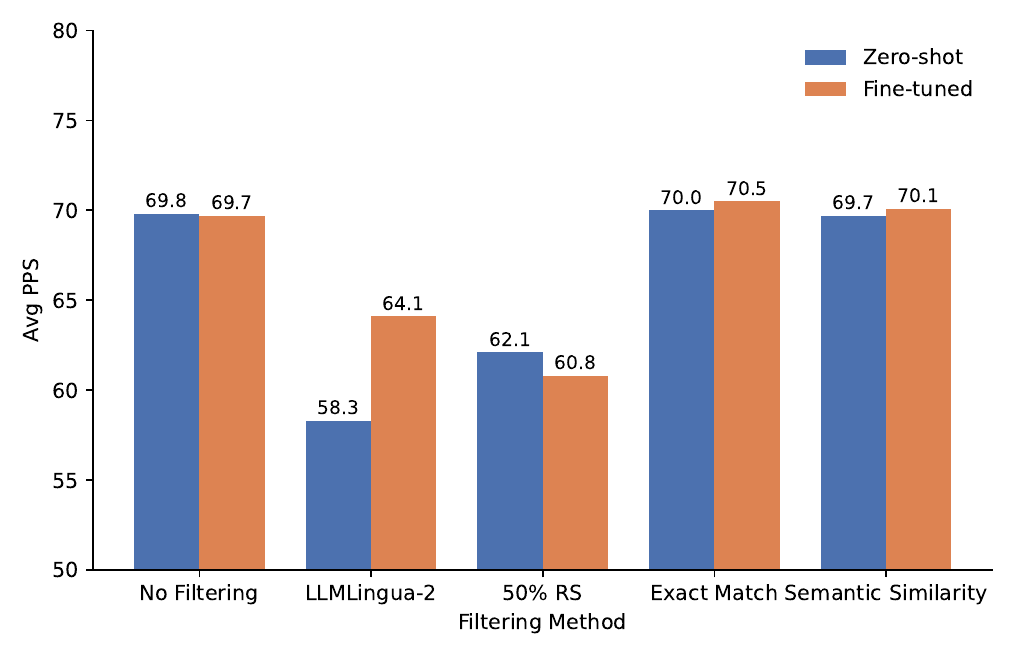}
    \caption{Impact of fine-tuning on Qwen3-4B average PPS metric across different filtering methods.}
    \label{fig:qwen3-4b-finetuning}
\end{figure}

Fine-tuning helps in our experiments only when the input has already been filtered (Figure~\ref{fig:qwen3-4b-finetuning}). Exact Match and Semantic Similarity improve to 70.5 and 70.1, both above the fine-tuned full-log baseline of 69.7, whereas training on the full or randomly sampled logs yields no improvement over the zero-shot setting. LLMLingua-2 gains the most in absolute terms, 5.8 points, but still trails every configuration except random sampling. Adaptation therefore compounds with filtering rather than substituting for it: removing redundant entries makes the remaining evidence easier for the model to learn to interpret.

\section{Discussion}

Our results indicate that persona inference benefits more from diverse behavioral evidence than from retaining every recorded event. Removing redundant entries reduced the memory footprint by up to 39\% and prefill time by more than half, while generally preserving or improving prediction quality.

The comparison among filtering methods further suggests that effective compression should preserve complete and distinctive events. Random sampling may discard infrequent but informative activities, while LLMLingua-2 operates at the token level without explicitly preserving event structure. In contrast, Exact Match and Semantic Similarity retain representative log entries. The strong performance of Exact Match indicates that a substantial portion of the redundancy is lexical, while Semantic Similarity additionally captures conceptually equivalent activities. Our fine-tuning results also suggest that filtering and model adaptation are complementary, particularly when the same filtering strategy is applied during training and evaluation.

For practical deployment, the memory reduction is more consequential than the latency reduction. Persona prediction can often be performed as a background task, making several seconds of additional prefill latency tolerable. Memory constraints are less flexible: a configuration that exceeds the available memory budget may not be deployable at all. We therefore view reducing the memory required by long-context inference as the primary practical benefit of log filtering.

\section{Conclusion}
We introduced fine-grained persona prediction from 24-hour mobile activity logs, a synthetic benchmark of 1{,}600 persona--log pairs, and a semantic similarity-based filtering method that removes redundant entries while preserving distinct behavioral evidence. At a full day's length, the KV cache alone can exceed a quantized 4B model's weights, making on-device persona inference a memory problem before a latency one. Filtering reduces the memory footprint by up to 39\% and prefill time by more than half, while maintaining or improving prediction quality. Structure-aware log compression therefore shifts the memory--quality trade-off toward practical, privacy-conscious on-device personalization.

\section*{Limitations}

Our study has several limitations. First, deduplication removes frequency information, although repetition may itself be informative for attributes such as communication habits, exercise routines, and application preferences. Future methods could preserve counts, durations, or temporal statistics alongside representative entries.

Second, our evaluation relies on synthetic data. Generated logs may be more coherent and more closely aligned with their corresponding personas than real mobile activity, so the results should be interpreted as evidence from a controlled benchmark rather than as an estimate of real-world performance. Validation on consented real-world datasets, such as StudentLife~\citep{wang2014studentlife} would be valuable, but the StudentLife dataset has only comparatively limited detail in persona annotations.

Third, we consider only a single 24-hour observation window. Longer or repeated observation periods may capture more stable routines and reduce sensitivity to atypical daily behavior. In addition, the Persona Prediction Score may obscure errors in individual attributes. Future evaluation should therefore consider multi-day inference, attribute-level performance, uncertainty calibration, and unsupported predictions.

Finally, our memory analysis is analytical rather than measured on mobile hardware. It accounts for model weights and the KV cache, but excludes activation buffers, allocator fragmentation, and runtime overhead, so the reported values should be viewed as lower bounds. Similarly, prefill times are measured with vLLM on server-class GPUs and are intended for relative comparison across filtering methods rather than as estimates of on-device latency. Measured on-device profiling remains an important direction for future work.

The filtering stage also introduces its own resource requirements through the Qwen3-Embedding-0.6B model and the FAISS index. In our pipeline, however, the embedding model completes before the language model is loaded, so the two memory costs are not incurred simultaneously. The peak footprint is therefore determined by the larger of the filtering and persona-inference stages rather than by their sum.

\bibliography{custom}
\clearpage
\onecolumn
\appendix

\section{Data Generation Prompts}
\label{sec:prompts}
Figure~\ref{fig:prompt_persona} shows the prompt used to generate the synthetic personas. Each persona is then used to generate a corresponding smartphone activity timeline using the prompt in Figure~\ref{fig:prompt_activity}. Finally, the persona and activity timeline are jointly used to generate the smartphone logs following the prompt in Figure~\ref{fig:prompt_log}.

\begin{tcolorbox}[breakable, enhanced, width=\textwidth,colback=orange!3!white, colframe=red!70!black, title=\textbf{Data Generation Prompts - Personas}]
\footnotesize\texttt{Generate a UNIQUE persona for \{random\_persona\} that authentically reflects cultural context while avoiding stereotypes.}

\medskip 

\texttt{To ensure diversity within the same demographic:}

\texttt{1. **Occupation**: Select age-appropriate roles combining sector and specific workplace (e.g., "18-year-old Korea: High school student", "19-year-old US: Part-time Safari Tour Guide", "45-year-old German: Senior Automotive Engineer at BMW Headquarters"). Avoid clichés through unexpected combinations (e.g., "Brazilian AI Ethicist at São Paulo Tech Institute", "South Korean Marine Biologist Managing Busan Coastal Ecosystem Project"). Always include company/institution name.}

\texttt{2. **Lifestyle**: Vary urban/rural settings based on their living cities, income brackets, family structures. Ensure workplace location aligns with residence city (e.g., "Guro Digital Complex" for Seoul-based tech workers).}

\texttt{3. **Avoid stereotypes**: Replace overused tropes with contemporary realities (e.g., Iranian Female: Drone Photography Entrepreneur in Tehran, 52-year-old Australian Male: Indigenous Land Rights Lawyer in Darwin).}

\medskip 

\texttt{Cultural authenticity checklist:}

\texttt{- Name: Phonetically accurate but English-alphabet compatible}

\texttt{- Daily habits: Reflect real regional patterns (meal times, transportation)}

\texttt{- Tech usage: Localized apps/services (e.g., Yandex.Taxi in Kazakhstan, Grab in Malaysia)}

\texttt{- Workplace: Company names should reflect local corporate culture (chaebols in Korea, startups in Estonia)}

\medskip

\texttt{Output MUST use this JSON structure with detailed SPECIFICS:}


\texttt{\{}

\texttt{\quad "name": "[Culturally appropriate name in English]",}

\texttt{\quad "Demographics": \{}

\texttt{\qquad "Age": [Exact number],}

\texttt{\qquad "Gender": "[Male/Female/Non-binary]",}

\texttt{\qquad "Residence": "[City, Country]",}

\texttt{\qquad "Marital status": "[Married/Single/Divorced] + [Kids if any]"}

\texttt{\quad \},}

\texttt{\quad "Job \& Lifestyle": \{}

\texttt{\qquad "Occupation": "[Job title] at [Company/Institution](e.g., 'Vertical Farming Specialist at Sky Greens')",}

\texttt{\qquad "Work hours": "[Unique schedule]",}

\texttt{\qquad "Workplace location": "[Specific district/neighborhood]",}

\texttt{\qquad "Commute method": "[Local transportation mode, time from home]"}

\texttt{\quad \},}

\texttt{\quad "Daily Patterns": \{}

\texttt{\qquad "Sleep": "[Cultural sleep patterns]",}

\texttt{\qquad "Weekday routine": "[Country-specific routines]",}

\texttt{\qquad "Weekend routine": "[Authentic local activities]",}

\texttt{\qquad "Hobbies": "[Mix common/unique interests]"}

\texttt{\quad \},}

\texttt{\quad "Financial Behavior": \{}

\texttt{\qquad "Investment style": "[Region-typical approach]",}

\texttt{\qquad "Spending habits": "[Cultural consumption patterns]",}

\texttt{\qquad "Credit management": "[Local financial norms]"}

\texttt{\quad \},}

\texttt{\quad "Health \& Fitness": \{}

\texttt{\qquad "Exercise": "[Popular local exercises]",}

\texttt{\qquad "Health habits": "[Traditional/modern health practices]"}

\texttt{\quad \},}

\texttt{\quad "Social Activities": \{}

\texttt{\qquad "Key relationships": "[Family/social structures]",}

\texttt{\qquad "Community": "[Local community engagement]"}

\texttt{\quad \},}

\texttt{\quad "Technology Habits": \{}

\texttt{\qquad "Top 15 apps": "[Country-specific apps]",}

\texttt{\qquad "Devices usage pattern": "[Popular local brands/usage habits]"}

\texttt{\quad \},}

\texttt{\quad "Motivations \& Pain Points": \{}

\texttt{\qquad "Goals": "[Culturally relevant aspirations]",}

\texttt{\qquad "Stressors": "[Local socioeconomic challenges]"}

\texttt{\quad \}}  

\texttt{\}}

\end{tcolorbox}
\captionof{figure}{The prompt used to generate personas.}
\label{fig:prompt_persona}

\begin{tcolorbox}[breakable, enhanced, width=\textwidth,colback=orange!3!white, colframe=red!70!black, title=\textbf{Data Generation Prompts - Activity Timeline}]
\footnotesize\texttt{Generate a detailed smartphone activity timeline covering EXACTLY 24 hourly blocks from when the persona first fell asleep yesterday to when they begin sleeping today. Randomly select a day of the week, with an 80\% chance of it being a weekday (Monday to Friday) and a 20\% chance of it being a weekend day (Saturday or Sunday).}

\medskip

\texttt{Follow these rules:}

\medskip

\texttt{\quad 1. **Persona Implementation**:}

\texttt{\qquad - Reflect occupation/routine (e.g., work messages during office hours)}

\texttt{\qquad - Incorporate financial behaviors (e.g., stock app checks)}

\texttt{\qquad - Mirror tech habits from device/app preferences}

\texttt{\qquad - Align with social/community activities}

\medskip

\texttt{\quad 2. **Content Rules**:}

\texttt{\qquad - Cover EVERY hourly block (from wenting to sleep and today's sleep)}

\texttt{\qquad - Include ALL smartphone interactions if it needs:}

\texttt{\qquad • Notifications (specify app/sender and its contents if it has)}

\texttt{\qquad • Calls (duration/contact, or missed)}

\texttt{\qquad • Messages (received or sent)}

\texttt{\qquad • App usage (specific actions)}

\texttt{\qquad - Add contextual details in () where needed}

\texttt{\qquad - Maintain chronological sleep-wake cycle}

\texttt{\qquad - Use the 24-hour time format to express times. When referring to times, use the following conventions:}

\texttt{\qquad • Previous day: 'previous' (e.g. yesterday's time)}

\texttt{\qquad • Current day: 'today' (e.g. today's time)}

\texttt{\qquad • Next day: 'next' (e.g. tomorrow's time)}

\texttt{\qquad • For more detailed examples and guidance, please refer to the provided examples.}

\medskip

\texttt{\quad 3. **Output Requirements**:}

\texttt{\qquad - Use STRICT JSON format with this structure:}

\medskip

\texttt{Example Output:}

\texttt{\{}

\texttt{\quad "timeline": \{}

\texttt{\qquad "Thu Mar 06 22:01-23:00": ["Instagram DM from college friend: 'what r u doing'", "Set morning alarm", "Fell asleep at 22:45"],}

\texttt{\qquad "Thu Mar 06 23:01-24:00": ["sleep"],}

\texttt{\qquad "Fri Mar 07 00:01-01:00": ["sleep"],}

\texttt{\qquad ...}

\texttt{\qquad "Fri Mar 07 07:01-08:00": ["Woke to alarm at 7:15", "Checked KB Bank app", "KakaoTalk message from boss: 'did you finish?'"],}

\texttt{\qquad ...}

\texttt{\qquad "Fri Mar 07 22:01-23:00": ["YouTube Music playback: beattles"],}

\texttt{\qquad ...}

\texttt{\qquad "Sat Mar 08 01:01-02:00": ["Set thermostat via SmartThings", "Began sleeping at 22:50"]}

\texttt{\quad \}}

\texttt{\}}

\medskip

\texttt{Generate timeline for this persona:}

\texttt{\{persona\_dict\}}

\end{tcolorbox}

\captionof{figure}{The prompt used to generate smartphone activity timeline.}
\label{fig:prompt_activity}


\begin{tcolorbox}[breakable, enhanced, width=\textwidth,colback=orange!3!white, colframe=red!70!black, title=\textbf{Data Generation Prompts - Logs}]
\footnotesize\texttt{Task: Your task is to generate smartphone logging data ONLY for the \#TIME AND EVENT\# based on the provided persona.}
\medskip

\texttt{\# Importance:}

\texttt{The result should be written entirely in English.}

\texttt{The structure and range of information in the logging data **MUST BE** identical to the logging sample data.}

\texttt{Write the logging AS POSSIBLE AS DETAIL (RECOMMEND to write more than 50 logging lines for each \#TIME AND EVENT\#).}

\texttt{There MUST be no logging of user-initiated actions (ex., touch, screen on, location, keyboard, application open) after the user has fallen asleep based on the timestamp}

\texttt{Do not describe the logging event outside of the time in \#\#TIME AND EVENTS\#\#.}

\medskip

\texttt{ONLY WRITE THE LOGGING DATA without additional explanation in the output.}

\texttt{Before the user interacts with the smartphone, there must be a `screen status' log indicating that the screen is `turned on'.}

\medskip

\texttt{\# PERSONA :}

\texttt{\{persona\_str\}}

\medskip

\texttt{\# TIME AND EVENT:}

\texttt{\{time\_action\}}

\medskip

\texttt{\# Characteristics/Information of Logging Data:}

\texttt{\quad - Logging information is recorded **chronologically**.}

\texttt{\quad - Format the timestamp as `Day of the week Month Day Hour(24H):Minute:Second' (e.g. `Thu Mar 06 22:14:00').}

\texttt{\quad - Logs are generated when **specific events occur**, with the following details:}

\texttt{\quad - **Data format**: `timestamp | sensor | description`}

\texttt{\quad - **Sensor types and their recorded content**:}

\texttt{\quad - **Sleep**: Time when sleep begins or when getting out of bed after waking up.}

\texttt{\qquad - *A day's worth of logging data must include 3 events*: } 

\texttt{\qquad 1. Sleep start time from the previous day}

\texttt{\qquad 2. Wake-up time (exiting bed) on the current day}

\texttt{\qquad 3. Sleep start time on the current day}

\texttt{\qquad - **Screen Status**: Time when the smartphone screen is turned on/off.}

\texttt{\qquad  - *Events are recorded*:}

\texttt{\qquad  - Before starting to use the phone}

\texttt{\qquad  - After finishing phone usage}

\texttt{\qquad  - **Keyboard**: Exact Text INPUT via smartphone keyboard by the user.}

\texttt{\qquad  - **Applications**: Timestamp when a specific app is launched.}

\texttt{\qquad  - **Touch**: User screen touch interactions and its name.}

\texttt{\qquad  - **Notifications**: App name that sent the notification and the contents if it has.}

\texttt{\qquad  - **Location**:}

\texttt{\qquad  - **No repeated** logging if the location remains unchanged from the previous one.}

\texttt{\qquad  - Place type: Only `Home` or `Company`}

\texttt{\qquad - Distance from home}

\texttt{\qquad  - Movement status: either `stopping`, `walking`, `running`, or `riding vehicles`}

\texttt{\qquad  - **Message**: Timestamp of sent/received SMS messages with anonymized numbers (e.g., (person 16), (person 2)), allowing us to recognize frequent senders while maintaining anonymity.}

\texttt{\qquad - **Calls**: Timestamp of incoming/outgoing/missed calls with call durations with anonymized numbers (e.g., (person 16), (person 2)), allowing us to recognize frequent senders while maintaining anonymity.}

\medskip

\texttt{\quad \#\#\# Logging Data Sample}

\texttt{\quad *(Randomly extracted examples per sensor type. **Please refer only to the information and structure of the logging data**):}

\texttt{\qquad 0,Thu Mar 06 00:22:00 | sleep | started to sleep}

\texttt{\qquad 1,Thu Mar 06 07:14:00 | sleep | got out of bed, spend a total of 136.0 minutes in REM sleep, with a sleep efficiency of 97\%, and a total sleep duration of 412.0}

\texttt{\qquad 2,Thu Mar 06 07:32:25 | locations | somewhere in outside, 438 meter from home, riding vehicles}

\texttt{\qquad 3,Thu Mar 06 07:40:00 | activity | 12 meters and take 15 steps in the next 10 minutes}

\texttt{\qquad 4,Thu Mar 06 07:49:10 | locations | somewhere in outside, 2468 meter from home, walking}

\texttt{\qquad …}

\medskip

\texttt{\quad \#\#\# OUTPUT STRUCTURE SAMPLE}


\texttt{\quad \{}

\texttt{\qquad "0": "Thu Mar 06 23:03:15 | screen status | Phone screen turned on",}

\texttt{\qquad "1": "Thu Mar 06 23:03:17 | applications | Opened the app Clock",}     

\texttt{\qquad "2": "Thu Mar 06 23:33:50 | touch | Clicked [Back to library] in the app Kindle",}

\texttt{\qquad "3": "Thu Mar 06 23:34:02 | screen status | Phone screen turned off",}

\texttt{\qquad "4": "Thu Mar 06 23:40:15 | locations | Home, 0 meter from home,
stopping",}

\texttt{\qquad "5": "Thu Mar 06 23:40:00 | sleep | started to sleep",}

\texttt{\qquad ...}

\texttt{\quad \}}

\end{tcolorbox}

\captionof{figure}{The prompt used to generate smartphone logs.}
\label{fig:prompt_log}

\section{Persona Prediction Prompt}
\label{sec:persona_prediction_prompts}
Figure~\ref{fig:persona_predction_prompt} presents the prompt used in predicting the persona details from logs.

\begin{tcolorbox}[breakable, enhanced, width=\textwidth,colback=orange!3!white, colframe=red!70!black, title=\textbf{Persona Prediction Prompt}]

\footnotesize\texttt{Based on the analysis of logging data generated from a smartphone, we aim to understand the user. 
Please review the meaning of the logging data provided and use it to predict the persona.} 

\medskip

\texttt{Using the attached logging data, fill in the persona information in the provided JSON template:}

\medskip

\texttt{Logging information is recorded in chronological order.
Logs are recorded when specific events occur, with the following details:}

\medskip

\texttt{- The data format is: timestamp | sensor | description}

\texttt{- Sensor types and their recorded content are as follows:}

\texttt{\quad - sleep: Time when sleep begins or when getting out of bed after waking up.}

\texttt{\quad - screen status: Time when the smartphone screen is turned on or off.}

\texttt{\quad - applications: Time when a specific application is launched.}

\texttt{\quad - touch: Record of screen touches by the user.}

\texttt{\quad - keyboard: Text input via the smartphone keyboard.}

\texttt{\quad - notifications: App name and content of notifications sent by the smartphone.}

\texttt{\quad - location: Distance from home/work and user movement status (e.g., stationary, walking).}

\texttt{\quad - message: Time of sent/received SMS messages with specific phone numbers.}

\texttt{\quad - calls: Time of incoming, outgoing, or missed calls with specific numbers.}

\medskip

\texttt{\#\#\# Output Requirements:}

\texttt{- Detail Level: Match the specificity of the reference example (e.g., occupation titles, routines, residence).}

\texttt{- Unknown Values: Use "X" for attributes with insufficient log evidence.}

\medskip

\texttt{\#\# Reference Clarification:}

\texttt{- Category Matching: Only populate fields that **EXIST** in this reference structure (MUST BE).}

\texttt{- The provided example output illustrates the level of detail required (e.g., granular descriptors like "Winter indoor soccer league").}

\texttt{- Ignore the example’s specific values (e.g., "Edmonton, Canada"); they are unrelated to actual logs.}

\medskip

\texttt{\#\# Reference:}

\texttt{\{reference\}}

\medskip

\texttt{\#\# Logging Data:}

\texttt{\{logs\}}

\end{tcolorbox}

\captionof{figure}{The prompt used to predict persona from logs.}
\label{fig:persona_predction_prompt}

\section{Implementation Details}
\label{app:implementation}

\paragraph{Inference.} We run inference with vLLM version 0.19.0 at batch size 8 on NVIDIA A40 GPUs, distributed across two GPUs for the models up to 8B parameters and four for Qwen3-14B, Gemma-3 12B, and Ministral-3 14B. For a given model, the same vLLM version, batch size, and GPU allocation are used across all filtering methods, so that reported prefill times differ only in the input representation.

\paragraph{Fine-tuning.} We fine-tune Qwen3-4B with LoRA~\citep{hu2022lora} at rank 16, $\alpha = 32$, and dropout 0.1, training for 3 epochs at a learning rate of $10^{-6}$ with an effective batch size of 8. A separate model is trained for each input configuration. The training set, training budget, and hyperparameters are identical across configurations, so that only the filtering strategy varies.

\end{document}